\def\year{2022}\relax
\documentclass[letterpaper]{article} 
\usepackage{aaai22}  
\usepackage{times}  
\usepackage{helvet}  
\usepackage{courier}  
\usepackage[hyphens]{url}  
\usepackage{graphicx} 
\usepackage{natbib}  
\usepackage{caption} 
\DeclareCaptionStyle{ruled}{labelfont=normalfont,labelsep=colon,strut=off} 
\usepackage{algorithm}
\usepackage{algorithmic}

\usepackage{booktabs} 
\usepackage{multirow}
\usepackage{tabularx}
\usepackage{amsmath,amssymb}
\usepackage{subcaption}
\usepackage{bbding}
\usepackage{amsmath}
\usepackage{amssymb}
\usepackage[table,xcdraw]{xcolor}
\usepackage{here}
\usepackage{arydshln}
\usepackage{newfloat}
\usepackage{listings}
\floatstyle{ruled}
\newfloat{listing}{tb}{lst}{}
\floatname{listing}{Listing}
\title{Neural Network Module Decomposition and Recomposition}
\author {
    Hiroaki Kingetsu,\textsuperscript{\rm 1}
    Kenichi Kobayashi, \textsuperscript{\rm 1}
    Taiji Suzuki \textsuperscript{\rm 2}
}
\affiliations {
    \textsuperscript{\rm 1}    AI Laboratory, Fujitsu Limited\\
    Kanagawa, Japan \\
    \textsuperscript{\rm 2}    The University of Tokyo and RIKEN Center for Advanced Intelligence Project \\
    Tokyo, Japan \\
    h.kingetsu@fujitsu.com, kenichi@fujitsu.com, taiji@mist.i.u-tokyo.ac.jp
}

\usepackage{bibentry}

\begin{document}

\maketitle

\begin{abstract}
We propose a modularization method that decomposes a deep neural network (DNN) into small modules from a functionality perspective and recomposes them into a new model for some other task. Decomposed modules are expected to have the advantages of interpretability and verifiability due to their small size. In contrast to existing studies based on reusing models that involve retraining, such as a transfer learning model, the proposed method does not require retraining and has wide applicability as it can be easily combined with existing functional modules.
The proposed method extracts modules using weight masks and can be applied to arbitrary DNNs. Unlike existing studies, it requires no assumption about the network architecture. To extract modules, we designed a learning method and a loss function to maximize shared weights among modules. As a result, the extracted modules can be recomposed without a large increase in the size. 
We demonstrate that the proposed method can decompose and recompose DNNs with high compression ratio and high accuracy and is superior to the existing method through sharing weights between modules. 
\end{abstract}

\section{Introduction}
The success of machine learning using deep neural networks (DNNs) has led to the widespread adoption of machine learning in engineering. Therefore, rapid development of DNNs is highly anticipated, and reusing model techniques have been well studied. Many of the existing studies on reusing models such as transfer learning involve retraining, which is cost-intensive. Although smaller models are preferable in terms of interpretability and verifiability, well-reused models are typically large and hard to handle.
We propose a modularization method that decomposes a DNN into small modules from a functionality perspective and then uses them to recompose a new model for some other task. The proposed method does not require retraining and has wide applicability as it can be easily combined with existing functional modules.
For example, for problems where only a subset of all classes needs to be predicted, it is expected that the classification task can be solved with only a smaller network than the model trained on all classes. As an example, for a 10-class dataset and its subsets of a 5-class dataset, the parameter size of the DNN required for a 5-class classifier can be smaller than required for a 10-class classifier. The goal here is to decompose the 10-class classifier into subnetworks as DNN modules and then recompose those modules to quickly build a 5-class classifier without retraining. 

In this study, we consider the use of neural networks in environments where the subclasses to be classified frequently change. In this case, retraining of the model is required every time the task changes, which is a major obstacle in the application aspect. As an approach to solve this problem, we propose a novel method that extracts subnetworks specialized for single-class classification from a (large) trained model and combines these subnetworks for each task to classify an arbitrary subset of tasks.
We propose a methodology to immediately obtain a small subnetwork model $M_\mathrm{sub}$ capable of classifying an arbitrary set of subclasses from a neural network model $M$ (defined as the trained model) trained on the original set of classes.
This method involves the following two procedures:
(1) We decompose the DNN classifier model $M$ trained on $N$-classes dataset and extract subnetworks $M_\mathrm{sub}$ that can solve the binary classification of each class $c_{n}$  ($c_{n} \in C; |C|=N$).
(2) By combining the subnetworks, 
a composed modular network $M_{\mathrm{sub}(c_{i},c_{j},... )} (c_{i},c_{j} \in C) $ is constructed.
In our approach, we modularize the neural network by pruning edges using supermasks~\cite{supermask}. The supermask is a binary score matrix that indicates whether each edge of the network should be pruned or not. 

\begin{figure*}
    \centering
    \includegraphics[width=0.85\linewidth]{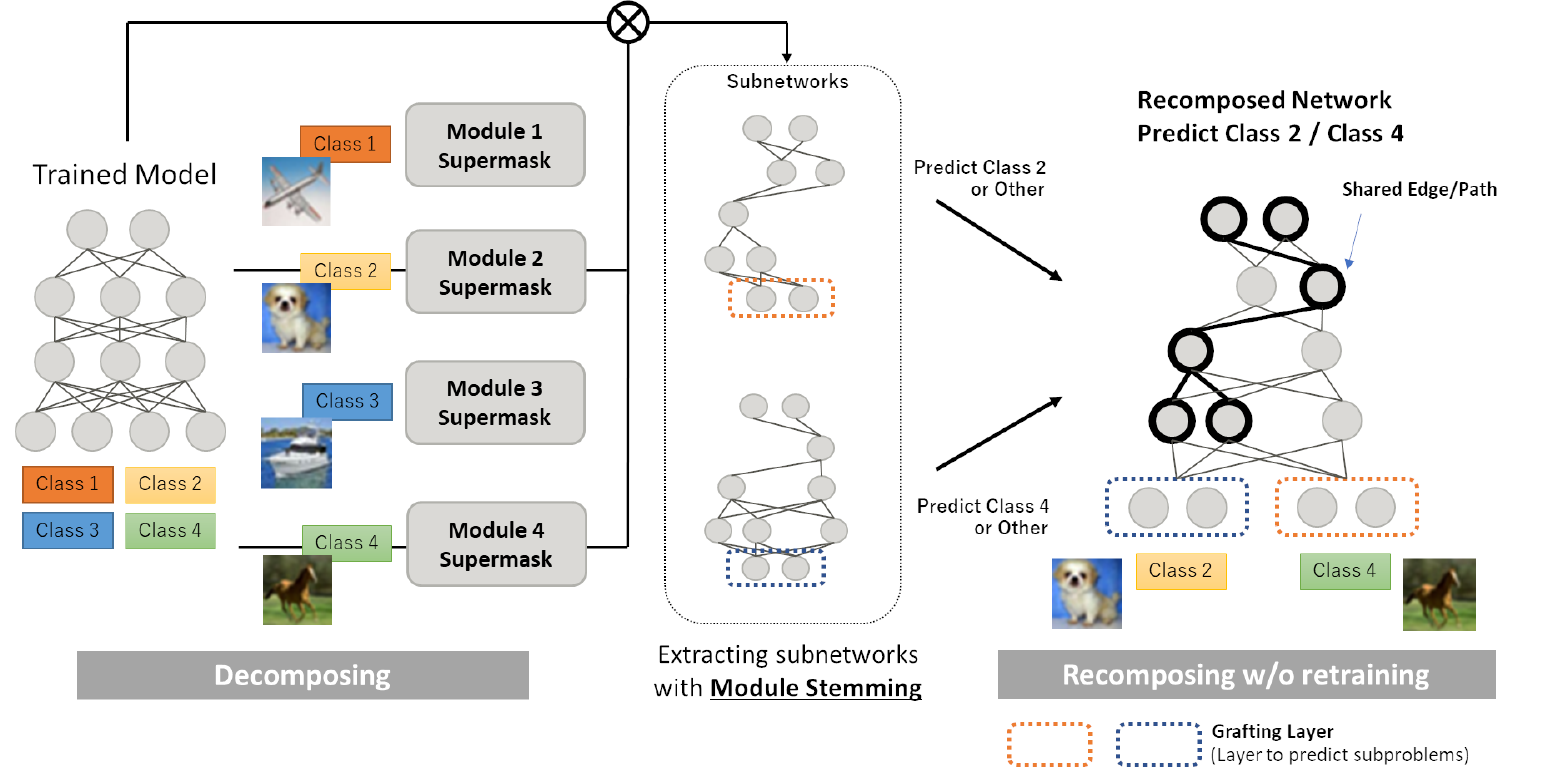}
    \caption{Overview of trained model decomposition and recomposition for the subtask.}
    \label{fig:overview}
\end{figure*}
Based on reports, even DNNs with random weights can be trained by only pruning branches with supermasks without changing the weights of the network~\cite{supermask,edge-popup}. In this study, we apply supermasks to the trained model to extract the subnetworks required for the classification of the subclass set from the trained model (Figure~\ref{fig:overview}). We also propose a new algorithm called module stemming that can construct a partial network with a minimum number of parameters by sharing the learning of the supermask among modules.
We demonstrate the effectiveness of our proposed method by comparing it with previous work using open datasets such as MNIST, Fashion-MNIST, SVHN, CIFAR-10, and CIFAR-100.

The contributions of this study are summarized below:
\begin{itemize}
    \item We applied supermasks to a trained DNN model and constructed a module network, which is a subnetwork of the trained model, by pruning networks. We showed that the trained model for multiclass can be decomposed into subnetworks to classify a single class. We also showed that the classification task of arbitrary subclass sets can be solved by a linear combination of these modules.
    \item We proposed a new method for learning a supermask that can be trained to prune similar edges between modules. By adding the consistency of the supermask score of each layer to the loss function, we show that the supermask can learn to remove dissimilar edges among modules and can classify them with a smaller parameter size during recomposing.
    \item We demonstrated the effectiveness of our proposed method for decomposing and recomposing modular networks using open datasets.
\end{itemize}

\section{Modular Neural Network}
\subsection{Why decompose a trained neural network into modules?}
The modularization of neural networks has been reported in many studies~\cite{auda1999modular, amer2019review, Compositionality}.
In these existing studies, training is performed at the architectural level to take on a modular structure.
The modular structure is said to be advantageous in the following ways:
\begin{itemize}
    \item The size of the network to be trained can be reduced, and the inference can be performed faster.
    \item The analysis of networks becomes easier.
    \item The effects of catastrophic forgetting~\cite{Catastrophic} in continual learning and incremental learning can be reduced.
\end{itemize}
Our approach is to extract and modularize subnetworks that can infer only one single class from a model trained on a general DNN model. These modular subnetworks can then be combined to address new tasks at the desired level.
The purpose of this study is not to use a special network architecture with a modular structure but to extract partial networks (subnetworks) with functional levels as modules that satisfy the above conditions from a large-scale trained model.
We handle these partial networks as module networks. These module networks are constructed as models that can classify a single class in binary to combine multiple modular networks and immediately obtain available models that can be classified for any subtasks.

\subsection{Required Properties}
\label{subsec:required_prperties}
In this study, we expect modular networks to have the following properties.
\begin{itemize}
    \item {\bf Decomposability.} Modular networks can be decomposed from a large network and have only a single classification capability. Each modular network should be smaller than the original network. Furthermore, the classification accuracy should be high.
    \item {\bf Recomposability.} Modules with a single classification capability can be combined with each other, and multiclass classification is possible by combining modules. Multiclass classification after recomposing should be performed with a high accuracy.
    \item {\bf Reusability /  Capability in small parameters.} Modules can be combined with only a simple calibration depending on the task. They can be reused without requiring relearning from scratch. Subnetwork $M_\mathrm{sub}$ should be represented with parameters as small as possible compared to the trained model $M$.
\end{itemize}
An approach that extracts from a large trained model is expected to be faster than learning a small model from scratch.
Methods such as transfer learning, which trains on large data and large models and fine-tunes them on a task, and self-supervised learning, which fits a problem dataset by pretraining on a large amount of unlabeled data, have been reported to perform promisingly. In the framework of extracting well-performing modular networks from such large models, we propose a method for extracting subnetworks from existing models through experiments.

\section{Related Research}
\subsection{Neural Network Decomposition}
Functional decomposition studies of neural networks (NNs) have been reported.
One of the major methods of NN decomposition is to consider the NN as a graph.
Clustering of network connections is considered a method to ensure the explanatory and transparent nature of the model~\cite{watanabe2018modular,SurprisinglyModular}.
Based on existing reports, it is possible to identify relationships between edges through low-rank approximation~\cite{low-rank}, weight masks~\cite{DifferentiableWeightMasks}, and ReLU activation~\cite{relu}. 
The goal of our study is modular decomposition for reuse, which means decomposition that can be recomposition.
Our approach is not to decompose the trained model into network layers (layer-wise) but to decompose it into classes to be predicted (class-wise). For example, in a class-wise decomposition, a modified backpropagation was developed by modularizing and decomposing the $N$-class problem into $N$ two-class problems~\cite{2class-AnandMMR95}. Anand et al. applied the proposed method to each module trained to distinguish a class $c_{n}$ from its complement $\overline{c_{n}}$.
Using a similar approach, Pan et al.~\cite{decomposing_modules} proposed a decomposition and recomposition method of NNs by removing their edges from the network relationships. However, they did not address convolutional layers, and their experiments were limited to simple datasets such as the MNIST dataset. Thus, the method is not applicable to more complex deep learning models. The studies on network decomposition are still in the preliminary experiment stage.


\subsection{Lottery Ticket Hypothesis and Supermask}
Fankle and Carbin~\cite{lottery_ticket} proposed the lottery hypothesis that the NN contains sparse subnetworks, and if these sparse subnetworks are initialized from scratch and trained, the same accuracy as the original network can be obtained. According to the experimental results of this study, the initial weights of the neural network contain subnetworks (called winning tickets) that are important for classification.
They found the winning tickets by iteratively shrinking the size of the NN, masking out the weights that have the lowest magnitude.
Zhou et al.~\cite{supermask} demonstrated that
winning tickets achieve better than random performance without training. They proposed an algorithm to identify a supermask, which is a subnetwork of a randomly initialized neural network that achieves high accuracy without training.
To improve the efficiency of learning and searching for supermasks, Ramanujan et al.~\cite{edge-popup} proposed the edge-popup algorithm using the score of a supermask. 
The application of a supermask to NNs with random weights shows that it is possible to superimpose a subnetwork capable of solving a specific task by any subnetwork without changing the weights of the original trained model.
Because the supermask does not change the weights of the original (randomly initialized) model, it is expected to prevent catastrophic forgetting, which is a problem in continuous learning. Accordingly, there has been a widespread research on the application of supermasks, especially in continuous learning.
Studies that have used supermasks to address the problem of continuous learning can be found in Piggyback~\cite{Piggyback} and Supsup~\cite{supsup}.
These studies are based on superimposing different tasks on a randomly initialized NN using a supermask. Using this approach, we can assume that a trained DNN for $N$-class problem contains subnetworks that can solve the binary classification of each class. By using this interesting property of NNs, we show experimentally that it is possible to extract a small subnetwork that can classify a single class.

\section{Methods}
\label{sec:Methods}

\begin{figure*}
    \centering
    \includegraphics[width=0.86\linewidth]{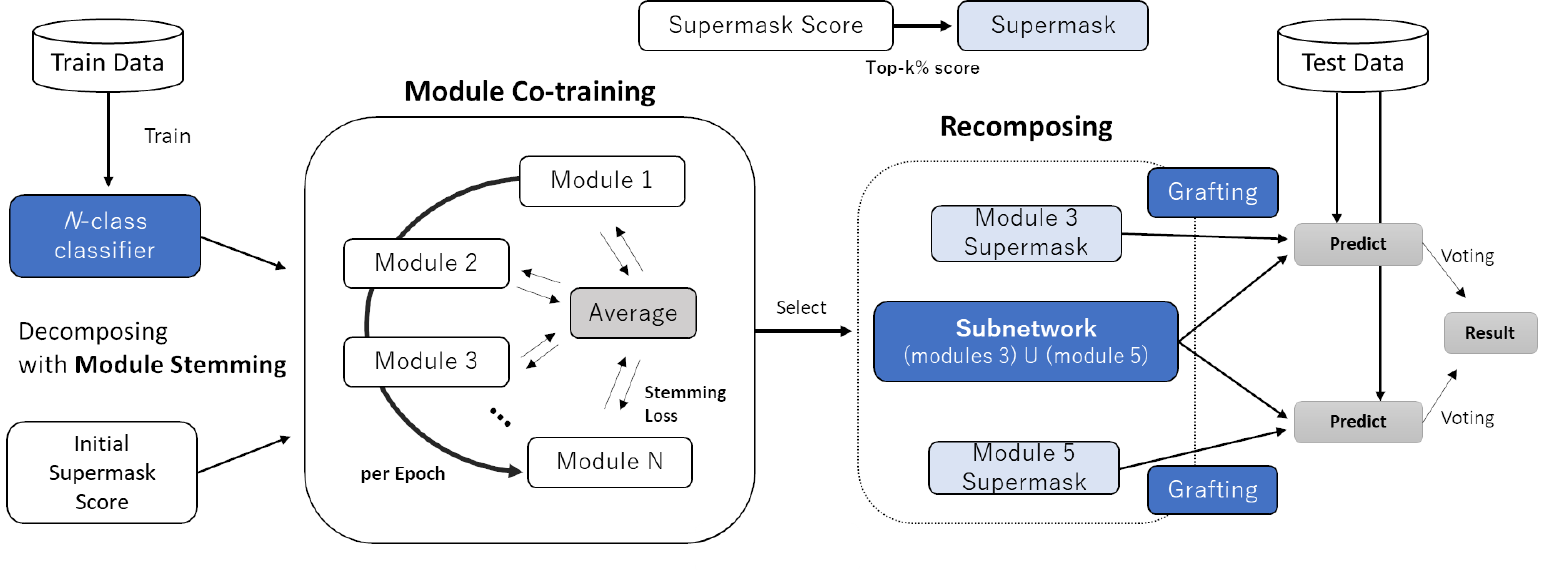}
    \caption{Overview of the model decomposition and recomposition flow for predicting the subtask.}
    \label{fig:process_model_composeing}
\end{figure*}

This study aims to decompose the trained model for $N$-class problems into subnetworks for a single problem prediction task and recompose NNs without retraining, as shown in Figure~\ref{fig:process_model_composeing}.
In this section, we describe the details of modularization comprising model decomposition using the supermask of the trained model, learning process of supermask, and recomposing a new NN of decomposed subnetworks for prediction use.


\subsection{Decomposing}
{\bf Training Mask}~~The module decomposition is achieved by applying supermasks, which indicates the branch pruning information, to the trained network. We follow the edge-popup algorithm~\cite{edge-popup} to calculate supermasks.
For clarity, we briefly describe the edge-popup algorithm for training supermasks. Let $x \in \mathbb{R}^{m}$ and $\hat{\mathbf{y}}=(W \odot \mathcal{M})^{\top} \mathbf{x}$ be the input and the output, where $W \in \mathbb{R}^{m \times n}$ are the trained model weights, $\mathcal{M} \in\{0,1\}^{m \times n}$ is the supermask, and $\odot$ indicates element-wise multiplication. Under fixed $W$ condition, the edge-popup algorithm learns a score matrix $\mathcal{S} \in \mathbb{R}_{+}^{m \times n}$ and computes the mask via $\mathcal{M}=h(\mathcal{S})$. The function $h$ sets the top $k \%$ of entries in $\mathcal{S}$ to 1 and the remaining to $0$. The edge-popup algorithm updates $\mathcal{S}$ on the backward pass optimized via stochastic gradient descent (SGD). The score $\mathcal{S}$ is initialized through Kaiming normal initialization~\cite{he2015delving} with rectified linear unit activation.
However, if the edge-popup algorithm is applied in a straightforward manner, the best performing edges to predict each different single class will remain, and the other edges will be pruned away. This condition leads to using different edges across modules to make predictions, which makes the model size very large when combined. Therefore, we propose a novel algorithm called {\bf module stemming} that uses common edges among modules and can be recomposed with few parameters.

\subsubsection{Grafting Layer}
An additional fully connected network layer called grafting layer is put after the last layer in the trained model (see Figure~\ref{fig:overview}). This layer changes from $N$-class prediction of the trained model output to the single-class prediction output of the modules. Single-class prediction estimates whether a class is a certain class or not. The weights in the grafting layer are trained by SGD with a low learning rate in the decomposition stage, but it is not retrained in the recomposition stage. Although it is possible to reuse only feature extraction from the trained model like transfer learning, we did not use this approach because the fully connected layer has the largest number of training parameters, and retraining this part of the model is not in line with the reusability objective of this study.
The grafting layer performs fine-tuning. It uses all $N$ logits of the $N$-class trained model (if not masked) to predict the single-class classification, thus allowing it to outperform the original $N$-class trained model for single-class classification. 

\subsubsection{Module Stemming}
For obtaining a subnetwork that can classify only the subtask, the parameter size of the module should be as small as possible. If the size of the parameters of the trained model is not significantly different from the model learned by recomposing the modules, then the trained model can be used as is, which is contrary to the purpose of this study. To obtain a small recomposed module (subnetwork), we can reduce the model size not only by reducing the network size of individual modules but also by making the supermask of each module as similar as possible and constraining the forward inference so that the same edges are used for inference. In this way, we can extract and build small subnetworks from the trained model. In this section, we discuss initial score sharing and stemming loss as ideas for extracting small network modules.

\begin{enumerate}
    \item \textbf{Initial Score Sharing}~~\\
    Pruning of a network depends on its initial parameter values. To construct modules consisting of similar edges, we take the simple approach of sharing the initial values of the super mask score between modules.
    \item \textbf{Stemming Loss}~~\\
For training supermask scores to build a module, we propose a loss function that regularizes the layer-wise score among the modules.
Consider linear layers that have the same neuron for simplicity, we define the following loss function $\mathcal{L}_{c_{n}}$ to learn a score matrix $\mathcal{S}_{c_{n}, \ell} \in \mathbb{R}_{+}^{m \times n}$ in a layer $\ell$ for the module $M_{\mathrm{sub}(c_{n})} $ by parameterized $W_{c_{n}}$ for each epoch while training with training batch data $\{x_{i}, y_{i}\}^{N_{d}}_{i=1}$:


\begin{multline}
\label{eq:loss}
\mathcal{L}_{c_{n}}= \frac{1}{N_{d}} \sum_{i=1}^{N_{d}}  \sigma \left(y_{i}, f(x_{i})) \right) + \\
\frac{\alpha}{L \times M \times N} \sum_{\ell}^{L} \sum_{m}^{M} \sum_{n}^{N} \left\| \mathcal{S}_{c_{n},\ell,m,n} - \frac{1}{N} \sum_{c_{n}=1}^{N} {\mathcal{S}^{\ast}_{c_{n},\ell,m,n}}\right\|_{p}^{p},
\end{multline}

where $\sigma(\cdot,\cdot)$ denotes cross-entropy, $\|\cdot\|_{p}$ denotes the $L_{p}$ norm, $\mathcal{S}^{\ast}$ denotes the scores in the previous epoch, and $\alpha$ denotes a hyperparameter.

\end{enumerate}
\begin{algorithm}[tb]
\caption{Module Co-training}
\label{alg:stemming}
\textbf{Input}: Training batch data $\{x_{i}, y_{i}\}^{N_{d}}_{i=1}$, trained model $M$ \\
\textbf{Parameter}: Parameter remaining $k$-\%, hyperparameter $\alpha$ \\
\textbf{Output}: Supermask scores $S_{c_{n}}$ for each class module
\begin{algorithmic}[1] 
\FOR{ $i = {1, ... N_{d}}$ }
\IF{first epoch}
\STATE Initialize score ${S_\mathrm{init}}$
\STATE Train supermasks $\mathcal{S}$ using ${S_\mathrm{init}}$ and $M$ without module stemming
\ELSE
\STATE Compute the average supermask scores ${S_\mathrm{avg}}$ of all modules in the previous epoch 
\FOR{each module}
\STATE Compute prediction output with $M$ and $\mathcal{S}_{c_{n}}$
\STATE Perform backpropagation with stemming loss (Eq.~\ref{eq:loss}) using ${S_\mathrm{avg}}$
\ENDFOR
\ENDIF
\ENDFOR
\end{algorithmic}
\end{algorithm}

\subsubsection{Module Co-training}
For constructing a base module for computing stemming loss, we tried two methods:
(A) We choose a single class arbitrarily and constructed the module without stemming loss. After that, we computed others with stemming loss.
(B) We used the average of the scores of all modules as the base.
Because method (A) does not have a methodology to select which class of modules to be based on, it was verified by brute force. However, as a result of trial and error, the accuracy of method (A) was not equal to or better than method (B).
In addition, method (A) is difficult to choose a good base class and a good choice depends heavily on the data set, and therefore we adopted method (B). We confirmed that efficient module stemming can be performed using method (B), called module co-training. The algorithm is shown in Alg.~\ref{alg:stemming}.


\subsection{Recomposing}
Given a subset of the $N$-class problem, the deployed model is a subnetwork of $N$-class problem classifier derived by obtaining the union of the supermasks for the classes in the subset. Therefore, if we can compute similar supermasks via module stemming and solve the problem using common edges, the size of deployed model becomes smaller. The deployed model makes predictions by sequentially applying supermasks and the grafting layer and then voting by a confidence score (e.g., max softmax score) without retraining, as shown in Figure~\ref{fig:process_model_composeing}.



\section{Experimental Setup}
\subsection{Dataset}
\begin{description}
\item {\bf MNIST}~~
The MNIST dataset~\cite{MNIST} is a well-known dataset used in many studies and consists of various handwritten digits (0-9) images. There are 60,000 training examples and 10,000 test examples, with an equal number of data for each class label.

\item {\bf Fashion-MNIST (F-MNIST)}~~
The F-MNIST dataset~\cite{f-mnist} has two-dimensional binary images from different 10 clothes.
As with the MNIST dataset, there are 60,000 training examples and 10,000 test examples, with an equal number of data for each class label.

\item {\bf CIFAR-10 / CIFAR-100}~~
The CIFAR-10 and CIFAR-100 training dataset~\cite{cifar10} consists of 50,000 images coming and the test set consists of 10k images from the same 10 classes and 100 classes, respectively. All images have a 32 × 32 resolution.

\item {\bf SVHN}~~
The Street View House Numbers (SVHN) dataset~\cite{SVHN} has 73,257 images in the training set, 26,032 images in the test set. All images contain 32 × 32 colored digit images. 
\end{description}
\subsection{Models}
For evaluation, we use four fully connected models ({\bf FC1, FC2, FC3, FC4}) that have 49 neurons and 1, 2, 3, and 4 hidden layers, respectively, following~\cite{decomposing_modules}. In addition, to verify the effect of convectional layer with batch normalization and dropout, we use {\bf VGG16}~\cite{VGG} for CIFAR-10, {\bf ResNet50}~\cite{resnet} for SVHN and {\bf WideResNet28-10}~\cite{wide-resnet} for CIFAR-100.
These models were trained on corresponding datasets with fixed epochs and SGD with momentum. WideResNet28-10 was trained by RandAugment~\cite{RandAugment}, one of the leading augmentation methods.
To evaluate the proposed algorithm, we compared the results in reference~\cite{decomposing_modules} as a baseline method. They proposed six methods, but we selected the TI-I, TI-SNE and CM-RIE methods that showed the better performance for remaining similarity edges between modules in their proposed method. Note that the methods proposed in reference~\cite{decomposing_modules} are not algorithms to increase the sharing of parameters between modules.
Because the method proposed in reference~\cite{decomposing_modules} is not applicable to models with convolutional layers, we compared them using FC models.


\subsection{Evaluation}
To evaluate the performance of the proposed method, we compared it with previous studies according to the evaluation criteria presented in this section to verify whether the constructed modules have the required properties.

\subsubsection{Decomposability}
\begin{itemize}
    \item Jaccard Index
\end{itemize}
We use the Jaccard Index (JI) to measure the similarities between the decomposed modules. If JI is 0, there is no shared edges between two modules. If it is 1, two modules are the same.
A higher JI indicates modules can utilize similar NN paths for inference and the recomposing model size is smaller.

\begin{itemize}
    \item Accuracy over the test dataset for modules
\end{itemize}
To measure the performance of the DNN model, we use accuracy metrics. When we utilize the decomposed modules in the recomposing stage, the prediction is based on the subnetwork, as shown in Figure~\ref{fig:process_model_composeing}.
After superimposing the supermask, we use the output of the grafting layer to 
predict labels via a voting method.
The computing superimposing of the supermask is very fast because it is a matrix computation.
As for voting, when we input the data and run each module, the positive output label with the highest confidence score among the modules is taken as the predicted label. Based on this, we calculate the accuracy of the test dataset.

\subsubsection{Recomposability}
\begin{itemize}
\item Accuracy for the recomposed networks
\end{itemize}
We evaluated the accuracy with which a model that combines multiple modules can predict a subtask.

\subsubsection{Reusability}
\begin{itemize}
\item Number of remaining parameters
\end{itemize}
We evaluate the number of parameters (RPs) required when the modules are recomposed using module stemming.
The smaller the number of parameters, the better. 
If it is possible to infer the class of instances while sharing edges between modules, the number of parameters needed to solve the subtasks can be reduced, thus reducing the model size. This can be evaluated by how many parameter percentages are needed to predict the subtasks based on the trained model.

\section{Results}
\label{sec:reseult}


\subsection{Preliminaries}
As a preliminary experiment, we evaluated the $L_{p}$-norm in Eq.~\ref{eq:loss}. As a result of comparing $p={1,2}$, although there was no significant change in the best accuracy or the number of parameters required, $p=1$ tends to change more slowly, which facilitates a more detailed evaluation of the accuracy and the number of parameters, and thus we experimented with $p=1$.

\subsection{Evaluation of Decomposability}

\begin{table*}[]
\centering
\resizebox{0.8\linewidth}{!}{%
\begin{tabular}{ll|cccccccc}
\hline
Dataset                 &     & MNIST         & MNIST         & MNIST         & MNIST         & F-MNIST       & F-MNIST       & F-MNIST       & F-MNIST       \\ \hline
Model                   &     & FC1           & FC2           & FC3           & FC4           & FC1           & FC2           & FC3           & FC4           \\ \hline
TI-I                    & JI  & 0.47          & 0.64          & 0.44          & 0.53          & \textbf{0.78} & 0.64          & 0.52          & 0.56          \\
                        & Acc & 94.91\%       & 96.83\%       & 69.19\%       & 96.44\%       & 85.82\%       & 87.58\%       & 77.55\%       & 87.51\%       \\
TI-SNE                    & JI  & 0.47          & 0.65          & 0.45          & 0.55          &  0.78        & 0.65          & 0.53          & 0.57          \\
                        & Acc & 94.91\%       & 96.83\%       & 96.30\%       & 96.79\%       & 85.82\%       & 87.58\%       & 87.09\%       & 87.79\%       \\
CM-RIE                    & JI  & 0.43          & 0.63          & 0.43          & 0.53          & 0.75            & 0.63       & 0.51          & 0.55          \\
                        & Acc & 94.90\%       & 96.82\%       & 96.33\%       & 96.75\%       & 85.85\%       & 87.56\%       & 87.10\%       & 87.95\%       \\  \hdashline
\textbf{Module Stemming (ours)} & JI  & \textbf{0.60} & \textbf{0.71} & \textbf{0.67} & \textbf{0.67} & 0.72          & \textbf{0.67} & \textbf{0.66} & \textbf{0.75} \\
                        & Acc & 96.47\%       & 95.92\%       & 96.61\%       & 96.75\%       & 84.63\%       & 85.44\%       & 86.30\%\     & 86.42\%       \\ \hline
\end{tabular}
}
\caption{Benchmarking average {\bf Jaccard Index (JI)} of each module and the average accuracy (Acc) for the recomposed model of all modules on each test dataset for each method ($k=10\%$). A higher JI implies better performance.
Because the existing method cannot be applied to convolutional layers, we tested only on the fully connected layer. The proposed method outperforms in most conditions. Although the module stemming does not pursue accuracy as it aims to constrain the edges to be selected, it is competitive with the existing method. }
\label{tab:acc_bench}
\end{table*}

Table~\ref{tab:acc_bench} shows the average accuracy of modules and recomposed module via decomposition of the trained model. In the MNIST and F-MNIST datasets, our method outperforms the conventional method in terms of JI of the recomposed module. Moreover, the accuracy is also higher than the baseline in most cases.

\subsection{Evaluation of Composability and Reusability}
\begin{figure*}
     \centering
     \begin{subfigure}[b]{0.28\textwidth}
         \centering
         \includegraphics[width=\linewidth]{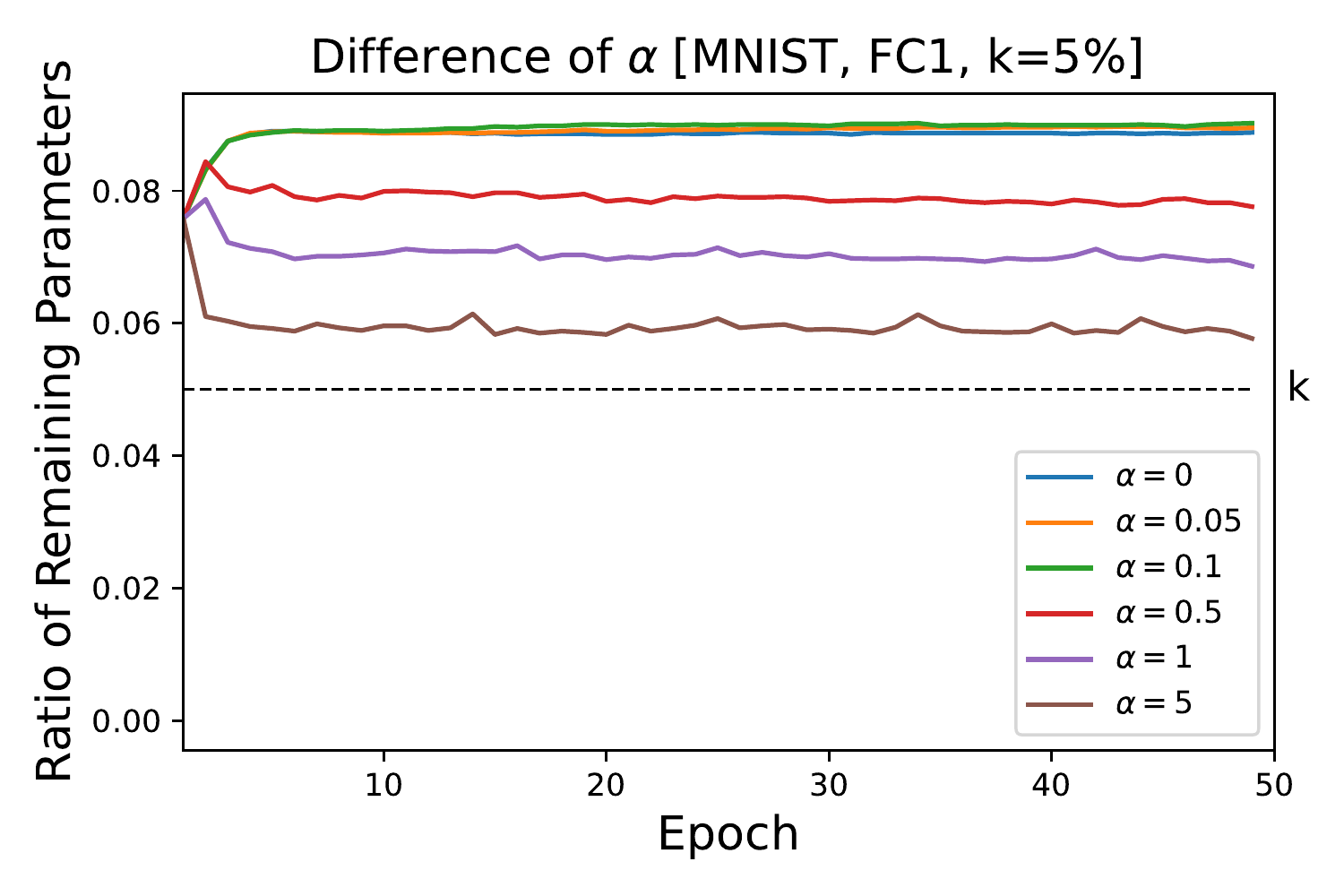}
     \end{subfigure}
     \begin{subfigure}[b]{0.28\textwidth}
         \centering
         \includegraphics[width=\linewidth]{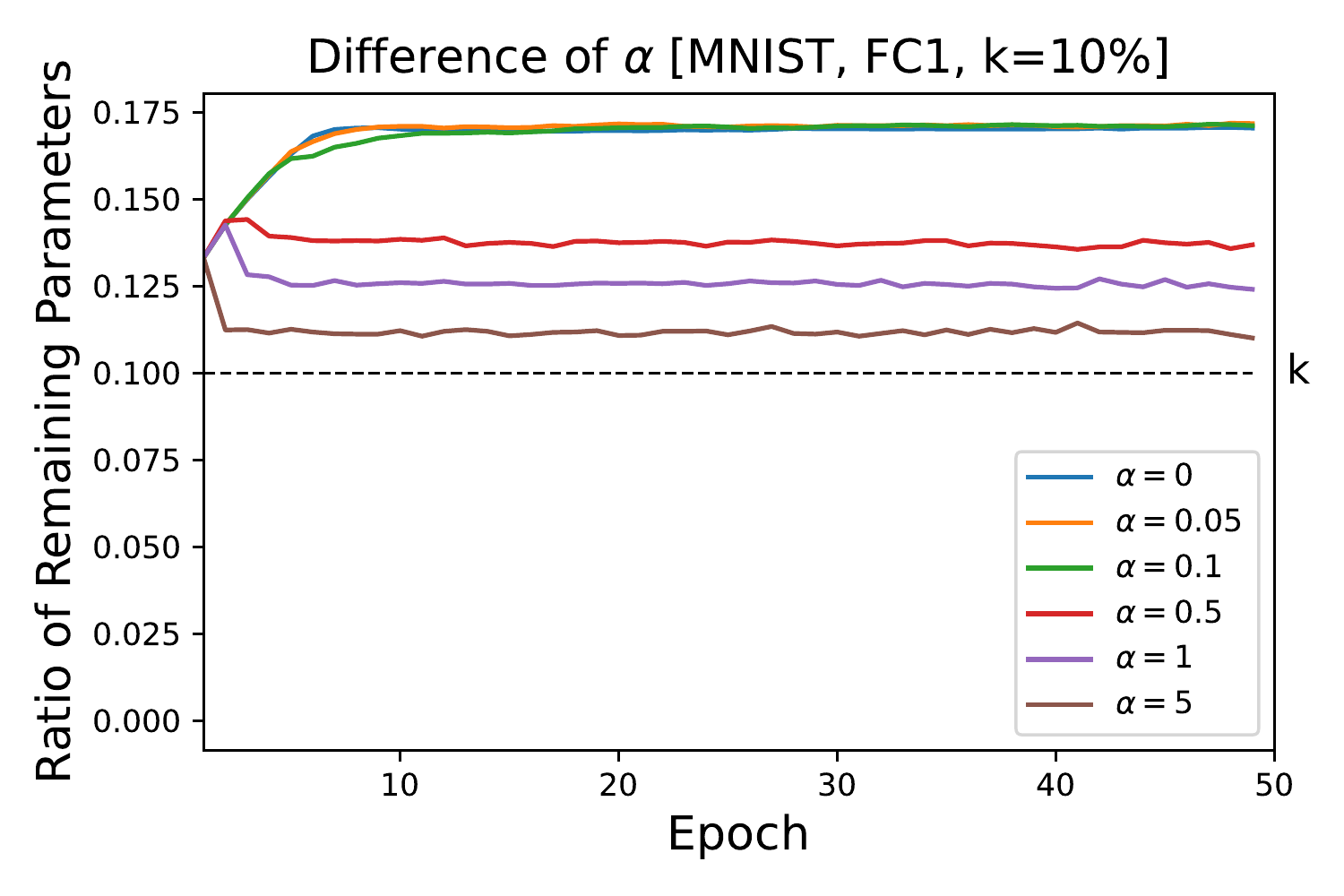}
     \end{subfigure}
         \begin{subfigure}[b]{0.28\textwidth}
         \centering
         \includegraphics[width=\linewidth]{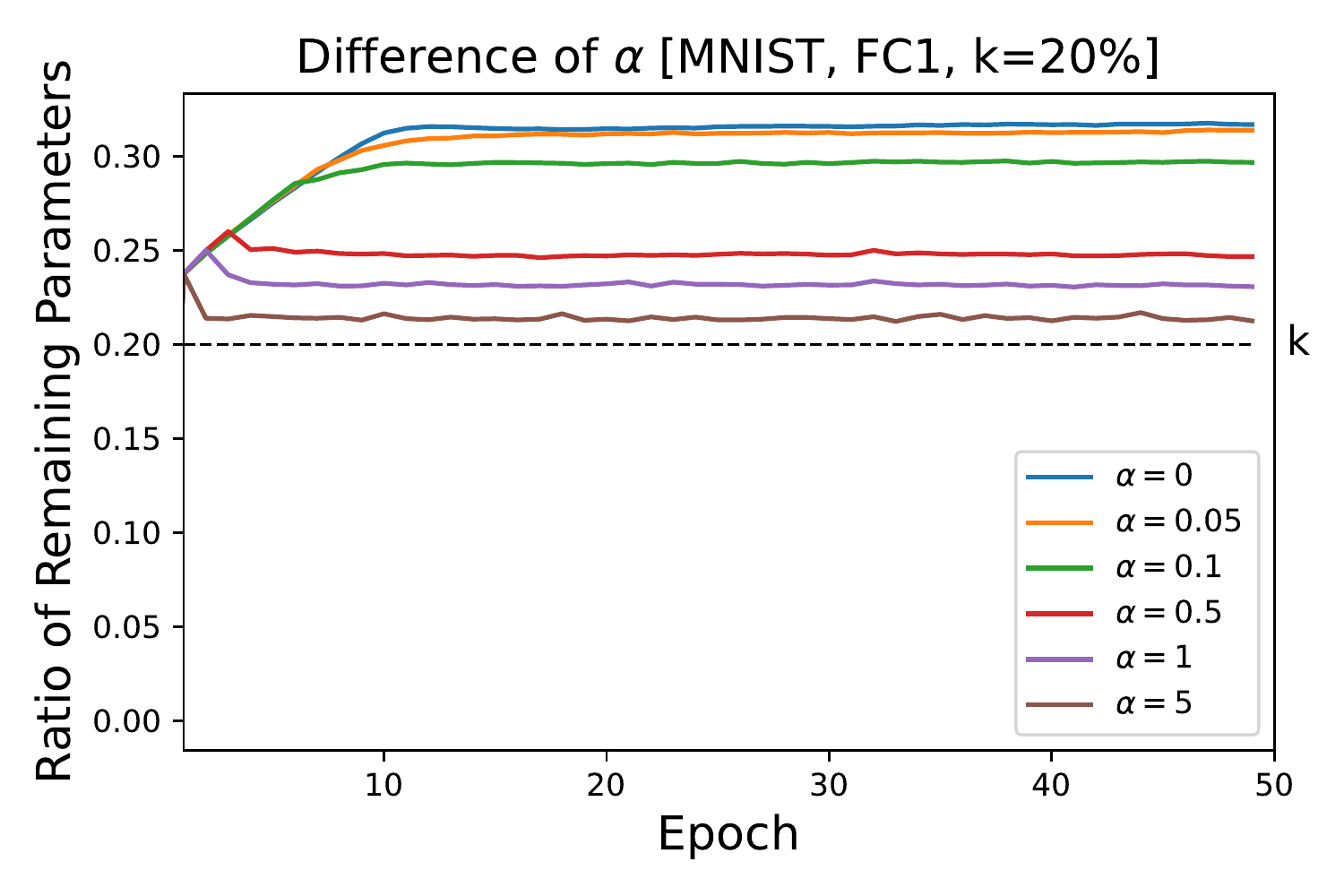}
     \end{subfigure}
     \hfill
     \begin{subfigure}[b]{0.28\textwidth}
         \centering
         \includegraphics[width=\linewidth]{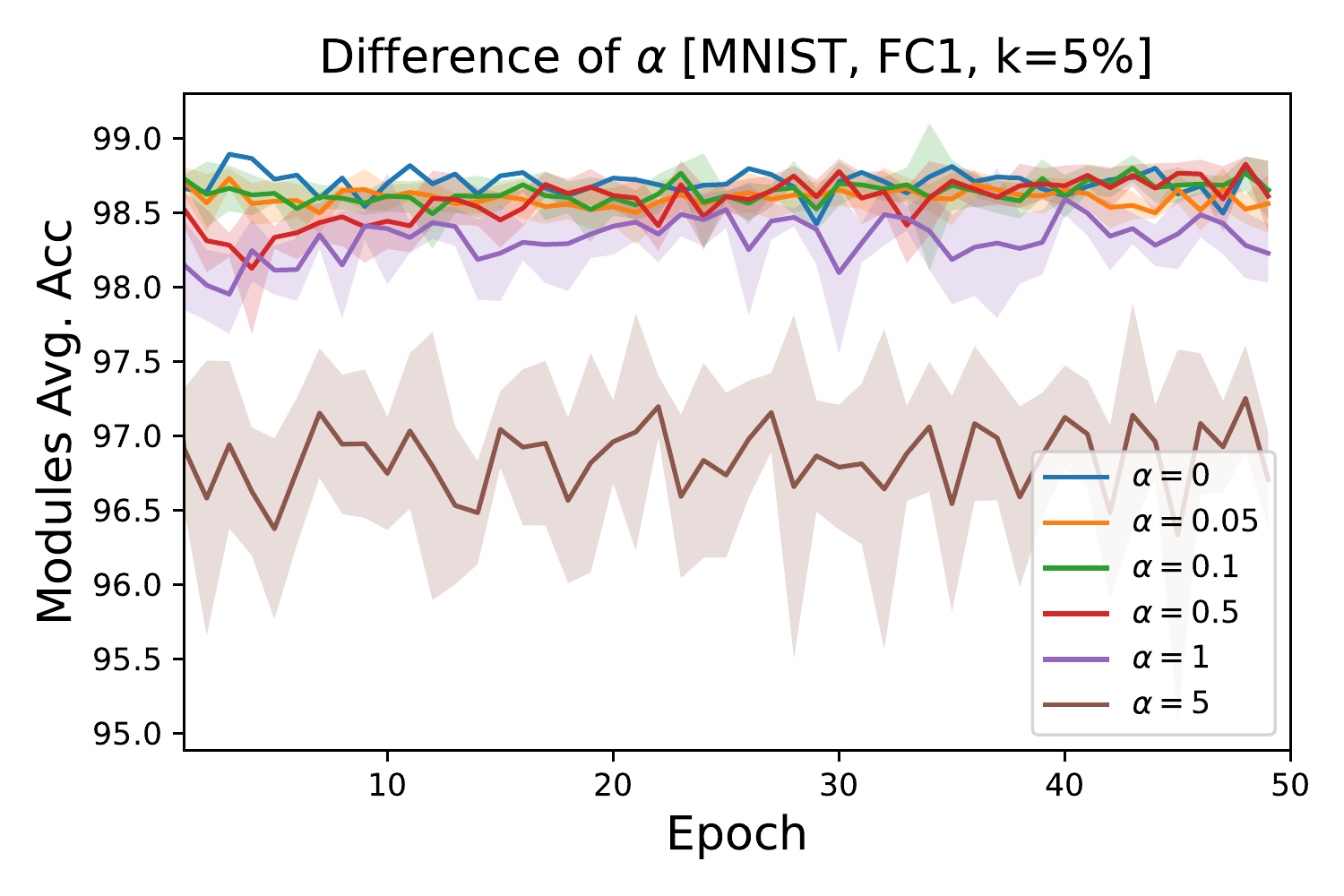}
     \end{subfigure}
     \begin{subfigure}[b]{0.28\textwidth}
         \centering
         \includegraphics[width=\linewidth]{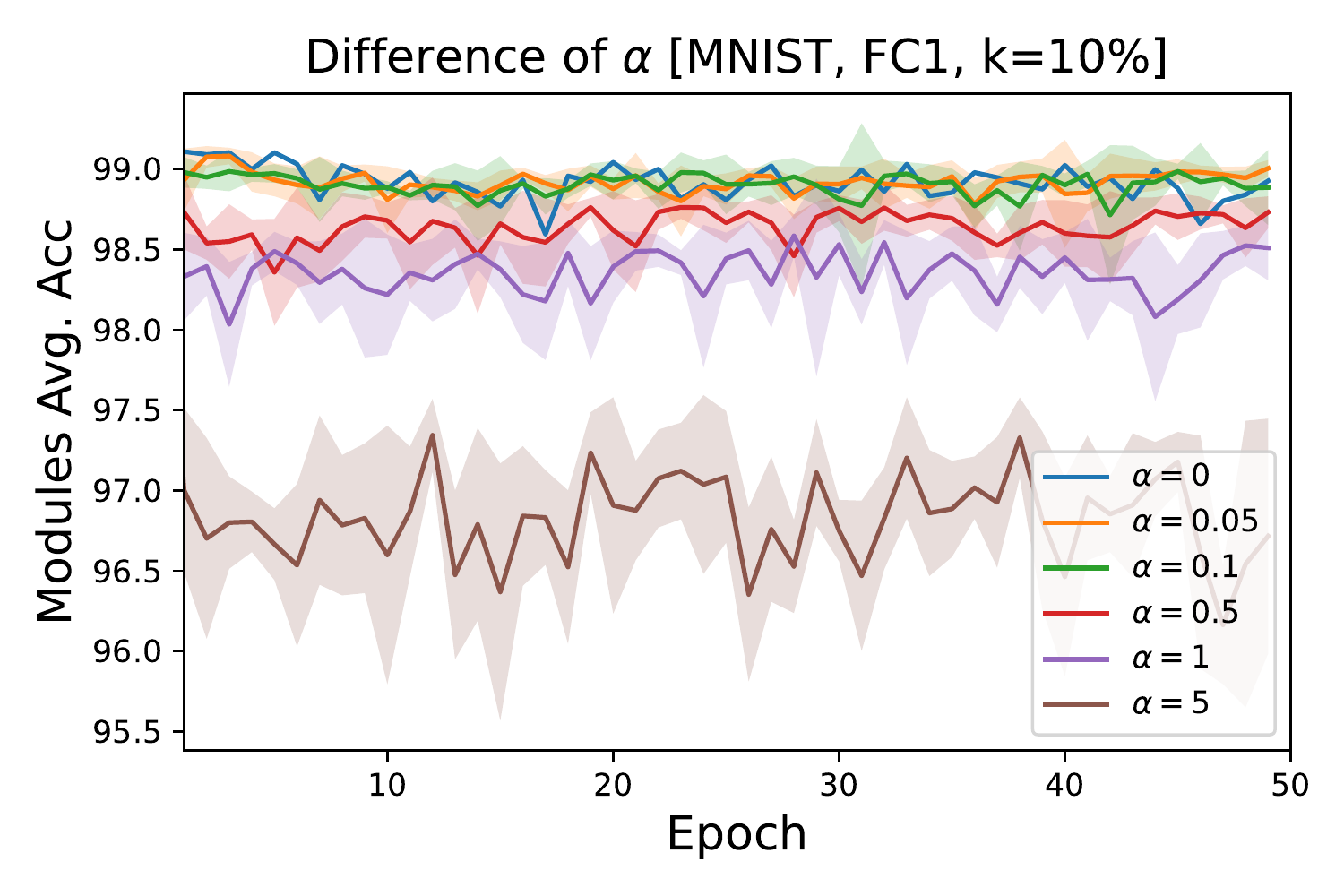}
     \end{subfigure}
     \begin{subfigure}[b]{0.28\textwidth}
         \centering
         \includegraphics[width=\linewidth]{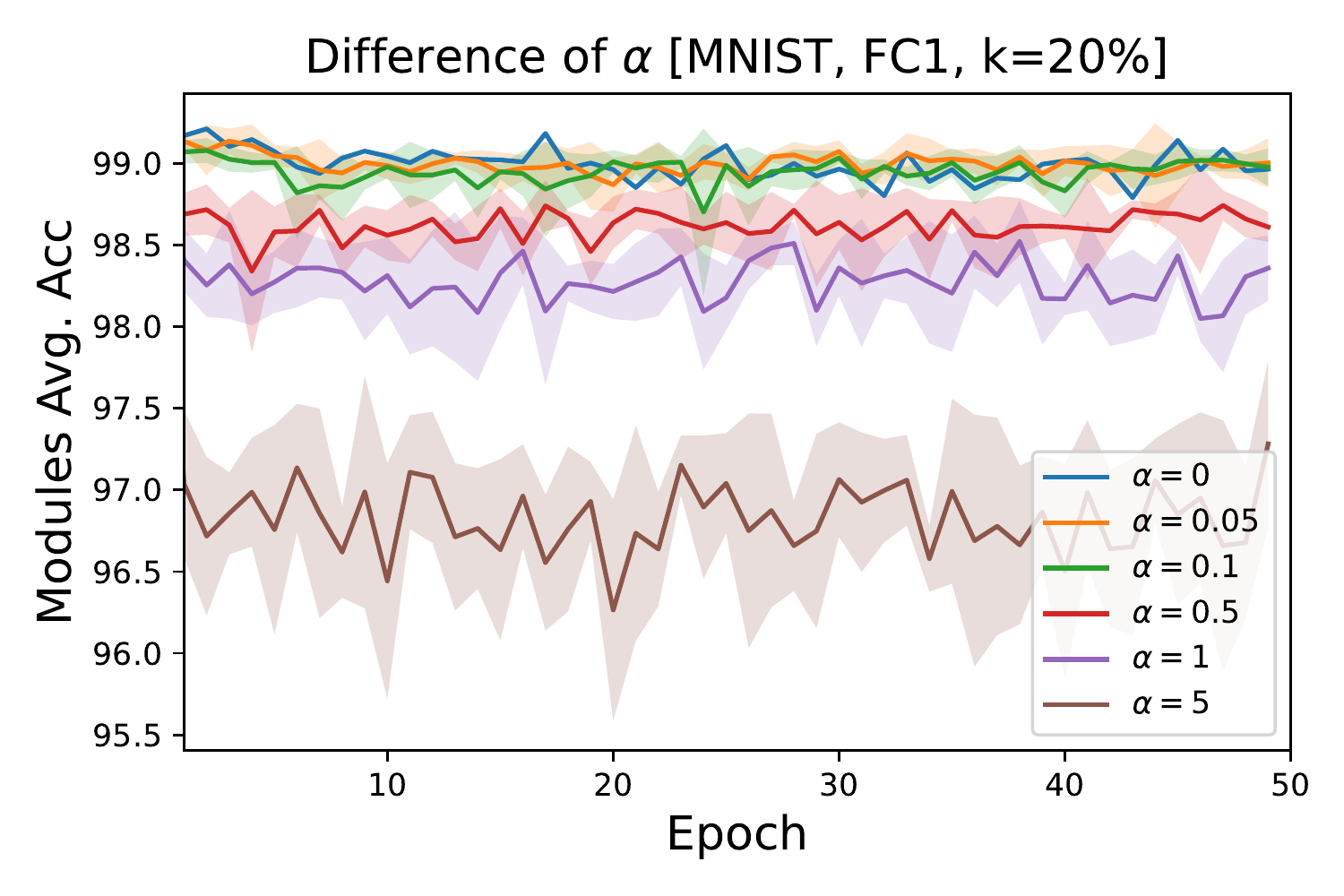}
     \end{subfigure}
    \caption{({\bf MNIST}) Top: Ratio of the average remaining parameters in the two-class module composition for each epoch when the hyperparameters are changed. Bottom: Average accuracy of the module. The dotted line (k) indicates the size of a single module.}
    \label{fig:avg_all_mnist}
    \end{figure*}
    \begin{figure*}
     \centering
     \begin{subfigure}[b]{\linewidth}
         \centering
         \includegraphics[width=0.28\linewidth]{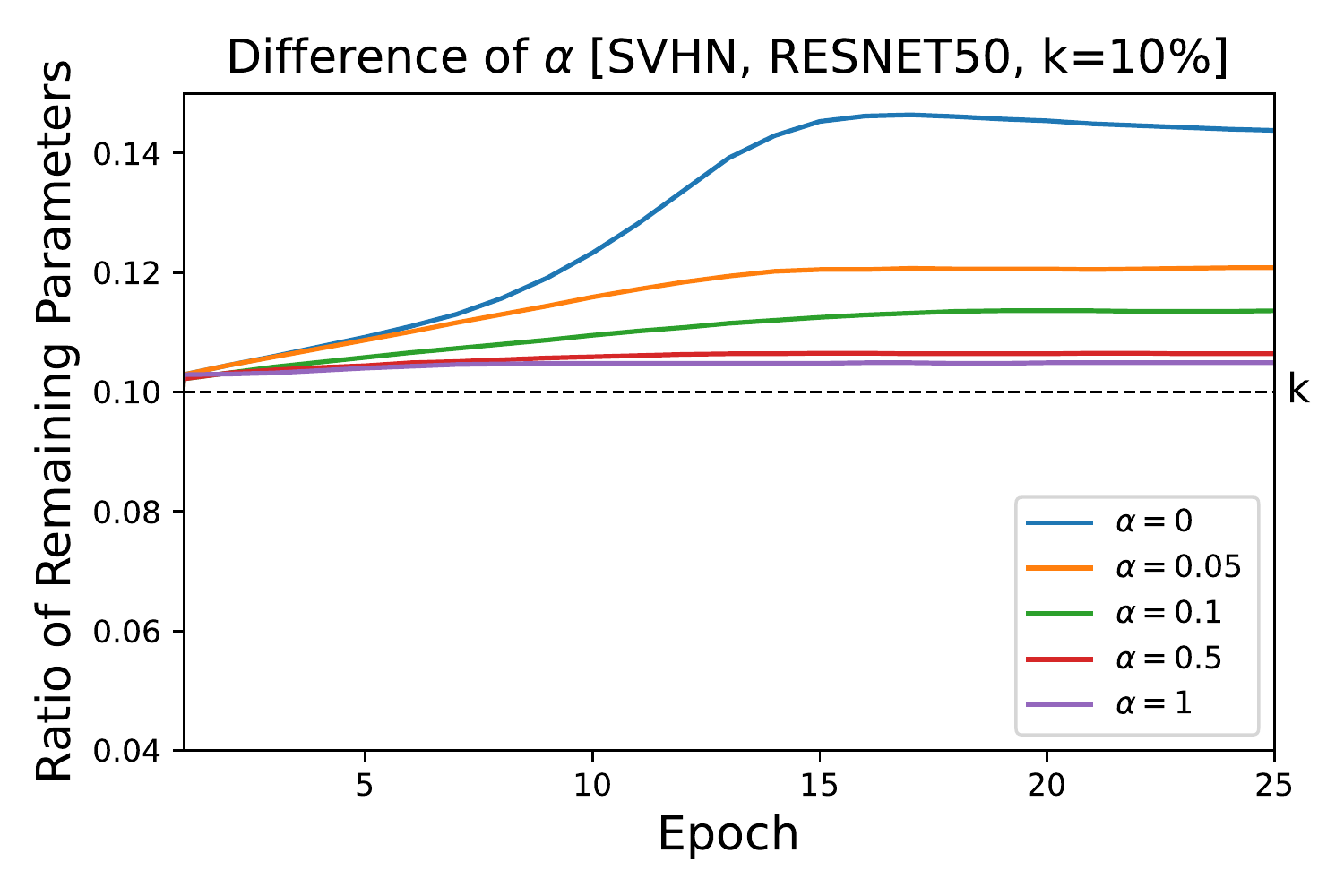}
         \includegraphics[width=0.28\linewidth]{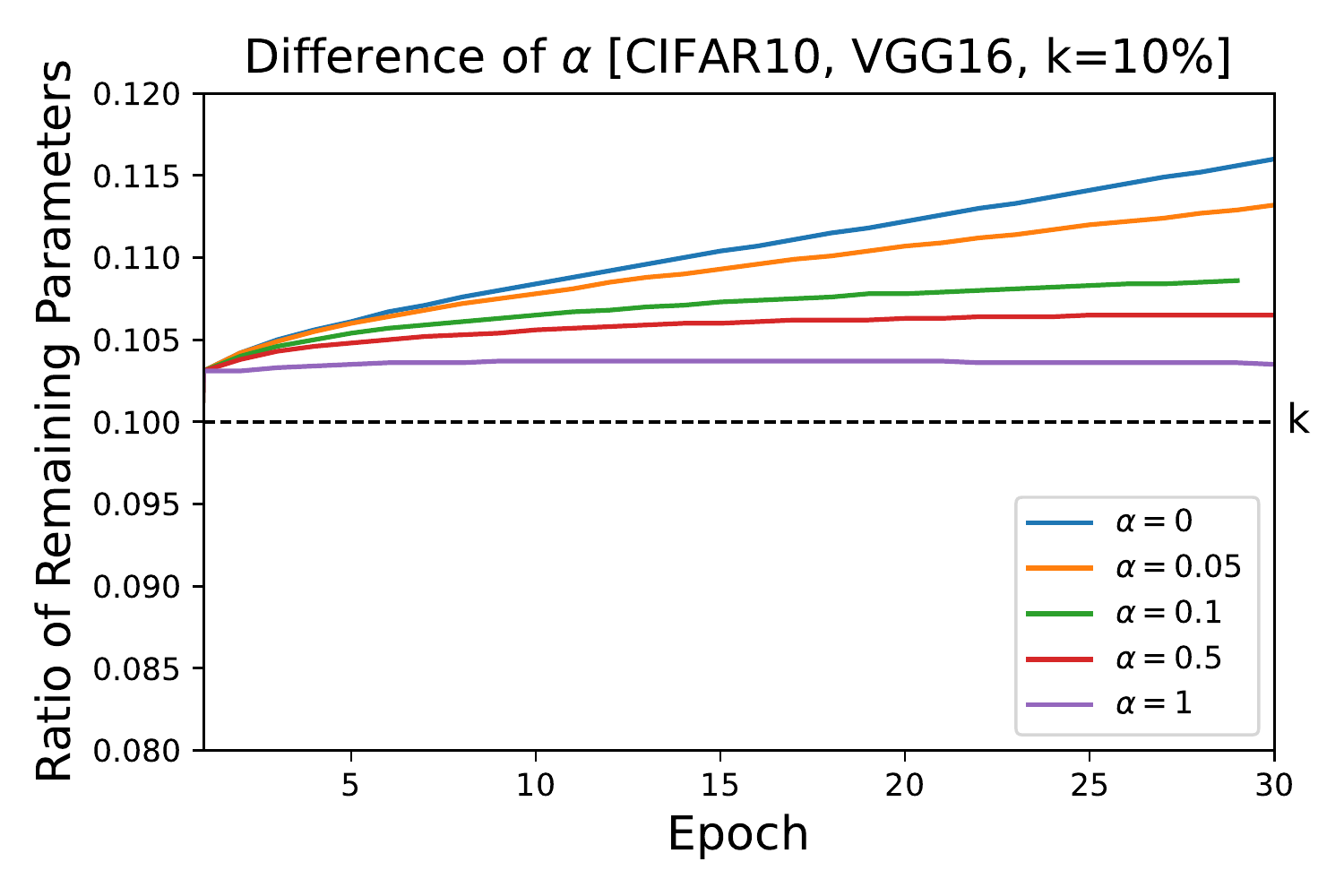}
         \includegraphics[width=0.28\linewidth]{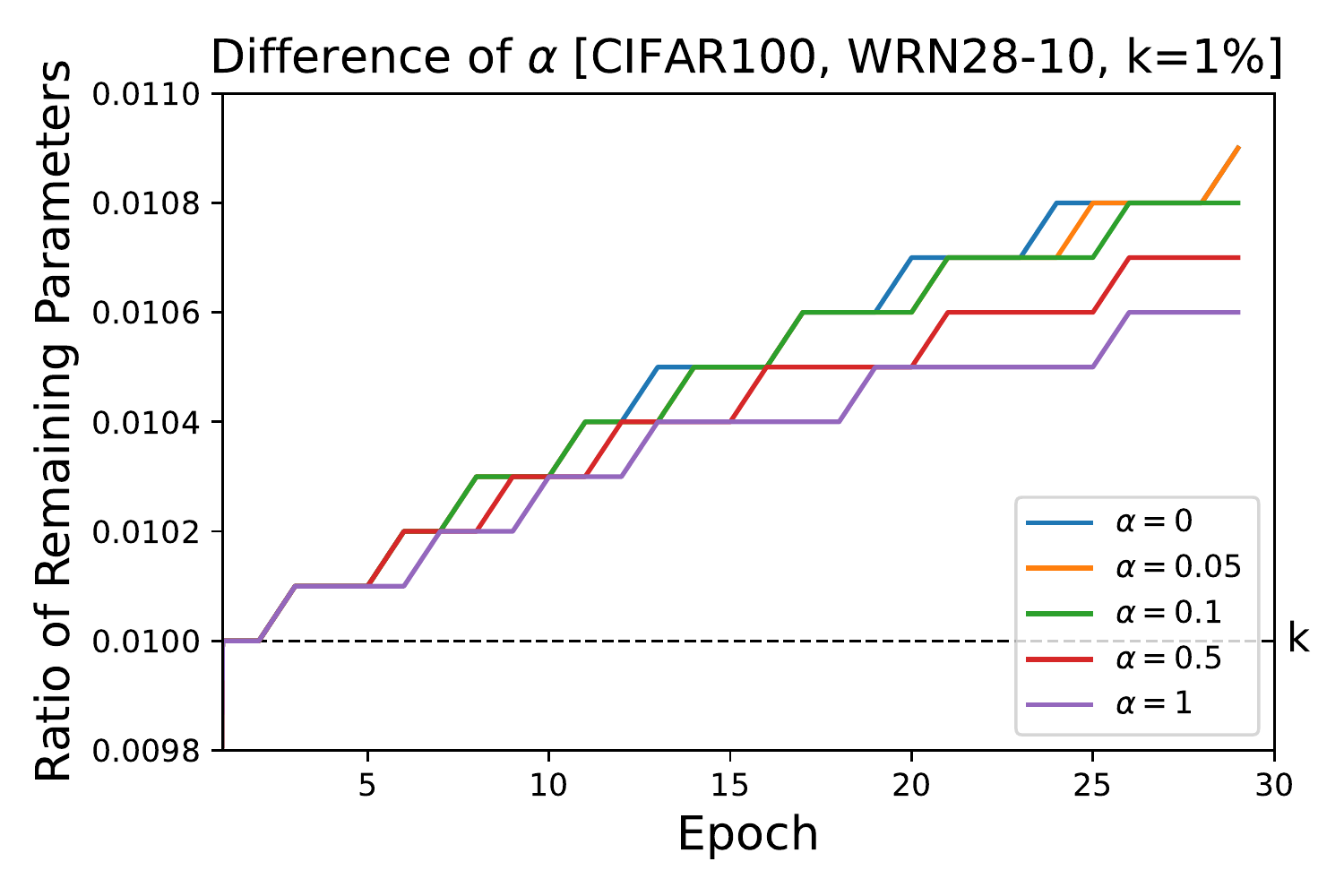}
     \end{subfigure}
     \begin{subfigure}[b]{\linewidth}
         \centering
         \includegraphics[width=0.28\linewidth]{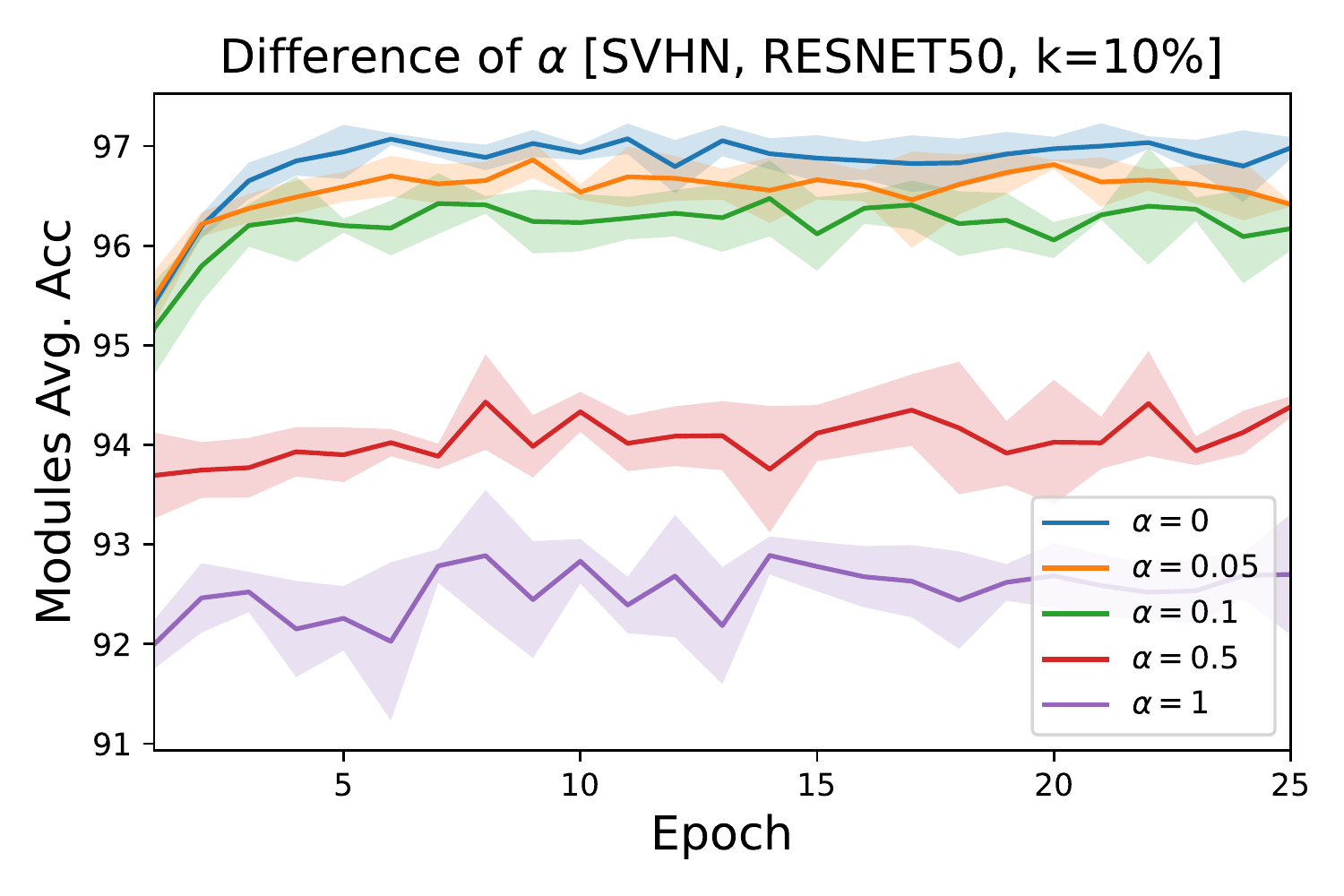}
         \includegraphics[width=0.28\linewidth]{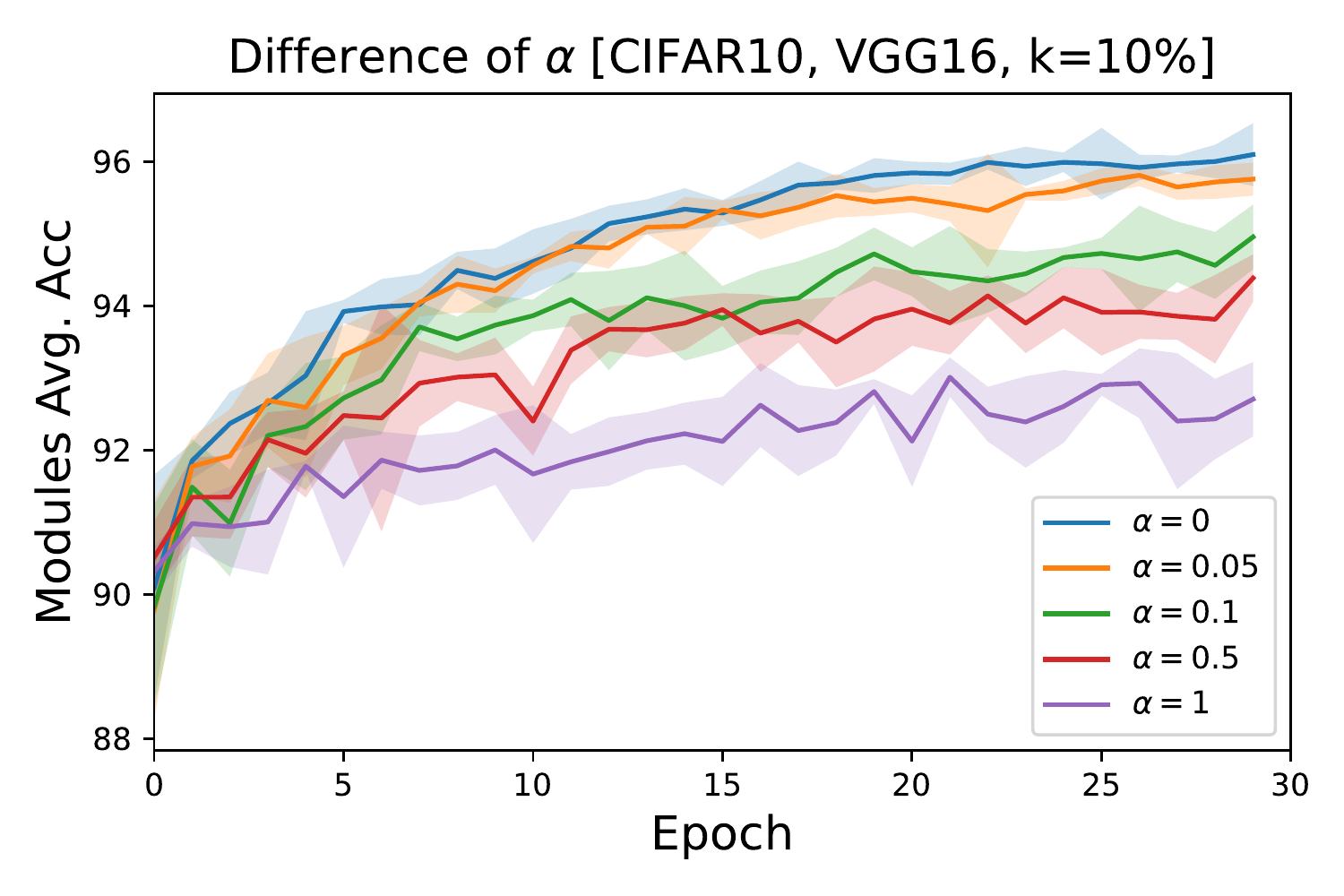}
         \includegraphics[width=0.28\linewidth]{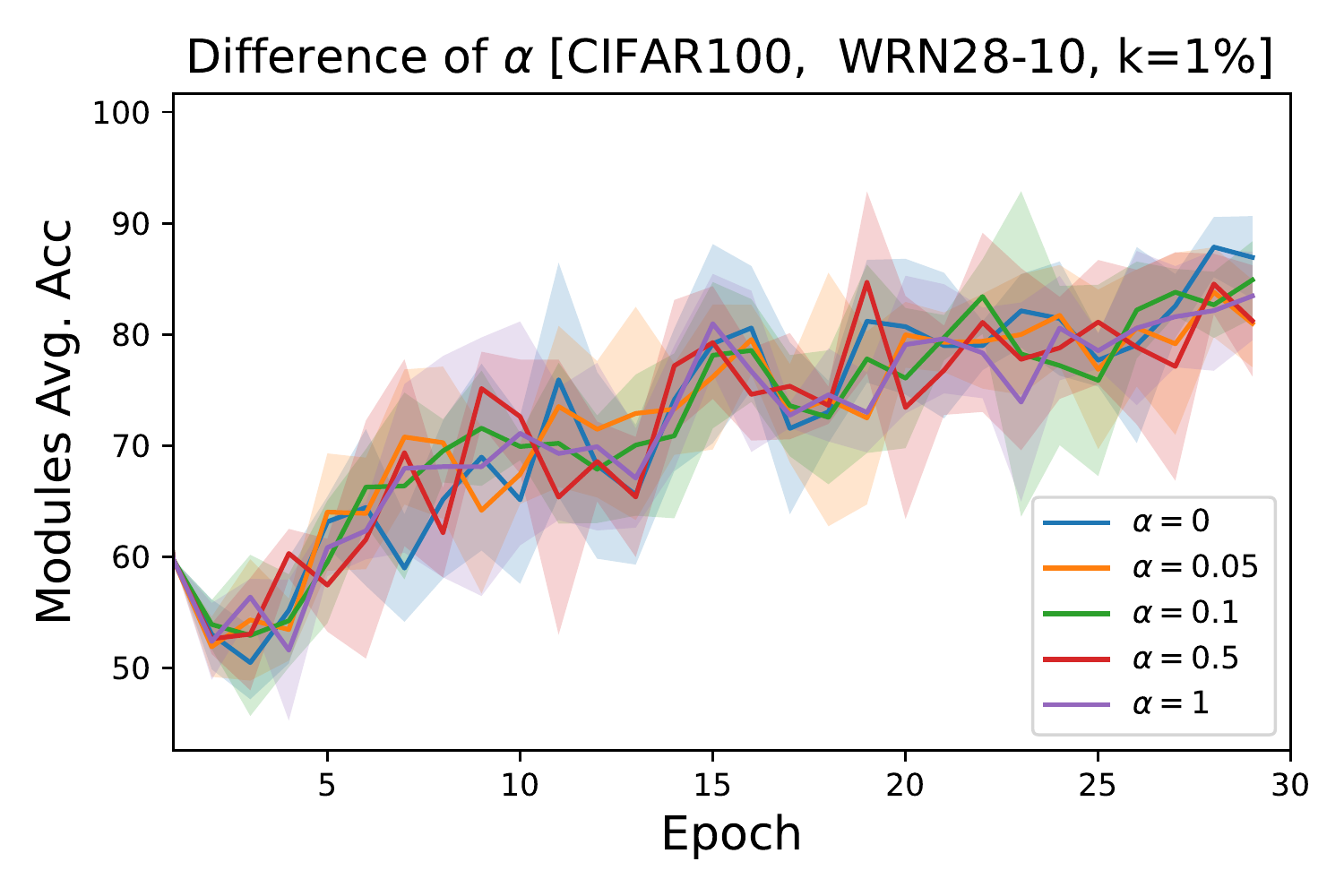}
     \end{subfigure}
     \hfill
    \caption{({\bf SVHN} with ResNet50 / {\bf CIFAR-10} with VGG16 / {\bf CIFAR-100} with WideResNet28-10) Top: Ratio of the average remaining parameters in the two-class module composition for each epoch when the hyperparameters are changed. Bottom: Average accuracy of the module. The dotted line (k) indicates the size of a single module.}
    \label{fig:avg_all_cifar}
\end{figure*}

The results in Figure~\ref{fig:avg_all_mnist} and Figure~\ref{fig:avg_all_cifar} show a much higher accuracy can be achieved and utilized as a small subnetwork by recomposing two modules (two-class problems). Each experiment is tested on five trials. Hyperparameter $\alpha$ is the coefficient for module stemming, and it the number of parameters required for recomposition is significantly reduced compared with the case where $\alpha=0$.
For example, when $k=10\%$ for model FC1 in Figure~\ref{fig:avg_all_mnist}, the ratio of RPs that is required to represent one module is $0.1$ for the original model. When two modules are composed, RPs is $0.1 + 0.1 = 0.2$ at worst, considering the two modules use exactly different edges for inference. In fact, it is close to $0.2$ when $\alpha = 0$ without module stemming. Moreover, due to module stemming, we can construct supermasks in which the modules are largely similar to each other after epochs converging to the RPs close to $0.5$.
Figure~\ref{fig:avg_all_cifar} shows the results for the CIFAR-10, CIFAR-100 and SVHN, which shows that the RPs due to module stemming is reduced, whereas the accuracy is not ensured for $\alpha=1$ compared to $\alpha=0$. This finding indicates that the number of parameters required for the CIFAR-10 / CIFAR-100 / SVHN classification is large in relation to the model size, and that the $\alpha$ for the module depends on the model size. In this experiment, although the value of $k$ is determined in accordance with baseline, it is necessary to determine $k$ and $\alpha$ depending on the model and dataset in terms of the performance. Similar to the model size compression method, we observed that the accuracy of the model decreased as the PRs increased. The proposed method is not suitable for environments that require significant class size because of the computational time required to build modules with module co-training. For actual use, it is necessary to verify the required recomposed model size and accuracy in advance. 

\subsection{Ablation study}
Table~\ref{tab:model_stemming} shows the comparison of the effects of the initial score sharing and stemming loss in the proposed methods. The results show that stemming loss could have a significant effect, and the low dependence of the score on the initial value was observed. Although initial score sharing has not produced a significant effect on performance, learning progresses in each module are expected to become more manageable.

\begin{table}[H]
\centering
\resizebox{\linewidth}{!}{%
\begin{tabular}{@{}cc|ccc@{}}
\toprule
Initial Score Sharing       & Stemming Loss     & Avg. Ratio of RPs & Best & Worst  \\ \midrule
           &            & 0.175                &    0.159                 &     0.185                    \\
\checkmark &            & 0.173                  &    0.159                  &    0.183                      \\
           & \checkmark  & 0.122                &     0.117                &   0.126                       \\
\checkmark & \checkmark & 0.122              &  0.116                  &   0.126                   \\ \bottomrule
\end{tabular}
}
\caption{Number of parameters required by the recomposition when building the module required to classify two subclasses at epoch 25 and $\alpha=1 ~(k=10\%)$ in MNIST with the FC1 model. {\bf Ratio of RPs} indicates the ratio of remaining parameters to solve a two-class classification task. A lower Avg. Ratio of RPs implies a good performance. The results show that the proposed method is effective in keeping the number of parameters low.}
\label{tab:model_stemming}
\end{table}

\subsection{Discussion}
Through experimental evaluation, we have confirmed that the proposed method can construct the module decomposition and merging of NNs for predicting classification problems with small parameters of the merging module for solving the subtasks and high accuracy. The conventional method is not applicable to networks that include CNNs, skip connection~\cite{resnet}, and average pooling~\cite{Global-Average-pooling}, and is limited to the fully connected layer.
The proposed algorithm, which performs pruning while retaining similar edges among modules, can be applied to widely used NNs using the proposed loss function based on differentiable masking.
As no training is required for recomposing, modules for predicting subtasks can be used immediately by applying masks.
As a limitation of our method, the module decomposition itself does not require a large amount of training time because it uses a trained model to cut the edges, but it consumes a computation time proportional to the number of classes. To decompose modules with a very large number of classes, it is necessary to use techniques such as distributed learning.

\section{Conclusions}
In this study, we address to extract the subnetworks required for the classification of the subclass set from the trained model. We propose an approach to decompose the existing trained model and modularize it. The proposed method employs weight masks to extract modules and is applicable to arbitrary DNNs. Moreover, it does not require any assumptions about the architecture of the network and can be applied immediately without the need for retraining. The proposed method has shown promising results, showing that it is able to extract similar edges across modules on several datasets. This allows us to reduce the model size when recomposing modules.
Future work includes a more detailed analysis of the edges in the decomposed model commonly used in module stemming, and further investigation of the conditions and model sizes under which stemming can work more efficiently.

\section{Acknowledgment}
TS was partially supported by JSPS KAKENHI (18H03201), Fujitsu Laboratories Ltd., and JST CREST.
\bibliographystyle{aaai22}
\bibliography{egbib}

\clearpage
\appendix


\section{Additional Experiments' Details}
\subsection{Training Parameter Details}


Table~\ref{tab:parameter} shows experiments' training parameters with SGD for building modules.
The details of our implementation code and training parameters essentially followed references~\cite{wide-resnet, supsup, decomposing_modules}. We confirmed that the accuracy of each model is almost equal to the accuracy presented in each reference.

\begin{table}[H]
\centering
\resizebox{\linewidth}{!}{%
\begin{tabular}{@{}llc@{}}
\toprule
\multicolumn{1}{c}{model} & \multicolumn{1}{c}{Parameters} & Value                          \\ \midrule
MNIST/F-MNIST/CIFAR10     & learning rate                  & 0.01                           \\
                          & weight decay                   & $5 \times 10^{-1}$             \\
                          & momentum(nesterov)             & 0.9                            \\
                          & batch size                     & 64                             \\
                          & epochs                         & 100                            \\ \midrule
CIFAR100                  & learning rate                  & 0.1 (stepLR)                   \\
                          & multi step LR decay            & 0.02 (epoch at 60,120,160) \\
                          & weight decay                   & $5 \times 10^{-1}$             \\
                          & momentum(nesterov)             & 0.9                            \\
                          & batch size                     & 256                            \\
                          & epochs                         & 200                            \\ \bottomrule
\end{tabular}%
}
\caption{Training Parameters}
\label{tab:parameter}
\end{table}




\subsection{Augmentation}
As shown in the previous study~\cite{edge-popup}, data augmentation can also improve the accuracy when using supermasks.
For CIFAR-10 and SVHN, we applied standard data augmentation (flipping and random crop).
We did not apply any data augmentation for MNIST and F-MNIST to compare with the previous study~\cite{decomposing_modules} under the same conditions. For the WideResNet28-10/CIFAR-100 models, we used RandAugment~\cite{RandAugment}. For the RandAugment parameters, we used in $N=2$ and $M=14$.

\subsection{Training Module Details}
For the build models via co-module training, we used parameters as shown in Table~\ref{tab:parameter_module}.
Each model decomposition was performed for five trials each, and the shaded error bars shown in Figure~\ref{fig:avg_all_mnist} and Figure~\ref{fig:avg_all_cifar}. indicate one standard deviation.
All models and modules were trained on eight NVIDIA Tesla V100 GPUs.

\begin{table}[H]
\centering
\resizebox{0.9\linewidth}{!}{%
\begin{tabular}{@{}lc@{}}
\toprule
\multicolumn{1}{c}{Parameters} & Value                            \\ \midrule
learning rate                  & 0.01 (WideResNet) / 0.1 (others) \\
weight decay                   & $5 \times 10^{-1}$               \\
momentum(nesterov)             & 0.9                              \\
batch size                     & 512 (WideResNet) / 32 (others)   \\ \bottomrule
\end{tabular}%
}
\caption{Decomposition Parameters}
\label{tab:parameter_module}
\end{table}

\section{Additional Analysis}

\subsection{Analysis of Hyperparameter $\alpha$}
Table~\ref{tab:acc-alpha} shows the results of the maximum test set accuracy of each module in MNIST for the comparison when the hyperparameter $\alpha$ is varied.
Under the condition of enough epochs, there is an upper bound of accuracy because there is no change of weights. In this experiment, under the condition of 50 epochs, there was no change in the best accuracy for 5 trials.
As shown in Table~\ref{tab:acc-alpha}, even when $\alpha > 0$, which has the effect of module stemming, the accuracy is sometimes best higher than $\alpha = 0$. This can be attributed to the inclusion of regularization effects in module stemming and the addition of a grafting layer.
Just as the accuracy does not necessarily increase with the number of layers or parameters, this indicates that the search for $\alpha$ is important as a hyperparameter.
Note that the results shown in Table~\ref{tab:acc-alpha} is the average accuracy of a single module (two-class problems), which is different from the accuracy after recomposition of 10 modules shown in Table~\ref{tab:acc_bench} (10-class problems).
\begin{table*}
\centering
\resizebox{0.75\linewidth}{!}{%
\begin{tabular}{@{}lccc|ccc|ccc|ccc@{}}
\toprule
\multicolumn{1}{l|}{}         & \multicolumn{3}{c|}{FC1} & \multicolumn{3}{c|}{FC2} & \multicolumn{3}{c|}{FC3} & \multicolumn{3}{c|}{FC4} \\ \cmidrule(l){2-13} 
\multicolumn{1}{l|}{}         & $\alpha$=0    & $\alpha$=0.05  & $\alpha$=0.1 & $\alpha$=0      & $\alpha$=0.05   & $\alpha$=0.1    & $\alpha$=0      & $\alpha$=0.05   & $\alpha$=0.1    & $\alpha$=0      & $\alpha$=0.05   & $\alpha$=0.1    \\ \midrule
\multicolumn{1}{l|}{module 0} & 99.5   & 99.63   & 99.58 & 99.66  & 99.68  & 99.68  & 99.67  & 99.67  & 99.67  & 99.43  & 99.68  & 99.67  \\
\multicolumn{1}{l|}{module 1} & 99.7   & 99.73   & 99.68 & 99.67  & 99.62  & 99.7   & 99.75  & 99.71  & 99.74  & 99.61  & 99.74  & 99.72  \\
\multicolumn{1}{l|}{module 2} & 99.06  & 98.97   & 98.9  & 99.19  & 99.18  & 99.14  & 99.2   & 99.21  & 99.23  & 98.87  & 99.12  & 99.14  \\
\multicolumn{1}{l|}{module 3} & 98.45  & 98.63   & 98.58 & 98.93  & 99.05  & 98.79  & 98.82  & 98.73  & 98.91  & 97.94  & 98.75  & 98.9   \\
\multicolumn{1}{l|}{module 4} & 99.24  & 99.22   & 99.2  & 99.22  & 99.30  & 99.28  & 99.36  & 99.26  & 99.16  & 98.47  & 99.06  & 99.16  \\
\multicolumn{1}{l|}{module 5} & 98.73  & 98.83   & 98.78 & 99.02  & 98.98  & 98.73  & 98.86  & 98.74  & 98.75  & 98.22  & 99.00  & 98.88  \\
\multicolumn{1}{l|}{module 6} & 99.44  & 99.45   & 99.45 & 99.48  & 99.49  & 99.41  & 99.47  & 99.37  & 99.38  & 99.19  & 99.42  & 99.44  \\
\multicolumn{1}{l|}{module 7} & 98.9   & 98.74   & 98.91 & 98.95  & 99.02  & 98.96  & 99.01  & 99.02  & 99.12  & 98.51  & 98.97  & 98.97  \\
\multicolumn{1}{l|}{module 8} & 98.06  & 98.18   & 98.54 & 98.39  & 98.31  & 98.51  & 98.42  & 98.59  & 98.5   & 97.67  & 98.66  & 98.62  \\
\multicolumn{1}{l|}{module 9} & 98.77  & 98.76   & 98.83 & 98.77  & 98.74  & 98.75  & 98.87  & 98.93  & 98.99  & 98.1   & 99.04  & 99.00  \\ \midrule
\multicolumn{1}{c}{avg}       & 98.99  & 99.01   & 99.05 & 99.13  & 99.14  & 99.10  & 99.14  & 99.12  & 99.15  & 98.60  & 99.14  & 99.15  \\ \bottomrule
\end{tabular}
}
\caption{The results of the maximum test set accuracy of each single module for the comparison when the hyperparameter~$\alpha$ is varied (MNIST ($k=10\%$)) }
\label{tab:acc-alpha}
\end{table*}

\subsection{Analysis of Modules Accuracy}
Related to Figure~\ref{fig:avg_all_mnist}, Table~\ref{tab:mono} shows how accurate each module is for classification on a dataset where each class is balanced with the other classes in a 1:1 ratio. The results in the table show that the accuracy is high only when the modules and subtasks are matched, but for different classes, binary classification is failed. 
This indicates that the modules can be successfully separated as subnetworks as single-class classification functions.


\begin{table*}[h]
\begin{subtable}[h]{\linewidth}
\centering
\resizebox{0.7\linewidth}{!}{%
\begin{tabular}{@{}ccccccccccc@{}}
\toprule
                              & class 0                       & class 1                       & class 2                       & class 3                       & class 4                       & class 5                       & class 6                       & class 7                       & class 8                       & class 9                       \\ \midrule
\multicolumn{1}{c|}{module 0} & \cellcolor[HTML]{C0C0C0}98.48 & 44.39                         & 44.64                         & 44.70                         & 44.46                         & 44.91                         & 45.12                         & 44.81                         & 44.83                         & 44.76                         \\
\multicolumn{1}{c|}{module 1} & 43.81                         & \cellcolor[HTML]{C0C0C0}98.38 & 43.99                         & 43.84                         & 43.80                         & 43.88                         & 43.93                         & 44.07                         & 44.13                         & 43.96                         \\
\multicolumn{1}{c|}{module 2} & 44.37                         & 44.22                         & \cellcolor[HTML]{C0C0C0}96.71 & 44.71                         & 44.50                         & 44.52                         & 44.76                         & 45.16                         & 44.64                         & 44.39                         \\
\multicolumn{1}{c|}{module 3} & 44.74                         & 44.92                         & 46.04                         & \cellcolor[HTML]{C0C0C0}92.43 & 44.79                         & 45.52                         & 44.72                         & 44.96                         & 46.16                         & 45.26                         \\
\multicolumn{1}{c|}{module 4} & 44.61                         & 44.51                         & 44.91                         & 44.59                         & \cellcolor[HTML]{C0C0C0}96.28 & 44.84                         & 45.02                         & 45.33                         & 44.75                         & 46.12                         \\
\multicolumn{1}{c|}{module 5} & 45.64                         & 45.54                         & 45.58                         & 46.19                         & 45.72                         & \cellcolor[HTML]{C0C0C0}92.42 & 45.95                         & 45.59                         & 46.03                         & 45.70                         \\
\multicolumn{1}{c|}{module 6} & 44.88                         & 44.67                         & 44.96                         & 44.74                         & 45.28                         & 45.13                         & \cellcolor[HTML]{C0C0C0}97.99 & 44.79                         & 44.86                         & 44.79                         \\
\multicolumn{1}{c|}{module 7} & 44.73                         & 44.71                         & 45.17                         & 44.99                         & 44.73                         & 44.86                         & 44.79                         & \cellcolor[HTML]{C0C0C0}94.21 & 44.94                         & 45.11                         \\
\multicolumn{1}{c|}{module 8} & 45.31                         & 45.45                         & 46.10                         & 45.68                         & 45.33                         & 45.82                         & 45.45                         & 45.43                         & \cellcolor[HTML]{C0C0C0}91.39 & 45.54                         \\
\multicolumn{1}{c|}{module 9} & 44.57                         & 44.45                         & 44.79                         & 44.77                         & 45.51                         & 44.89                         & 44.58                         & 45.31                         & 45.18                         & \cellcolor[HTML]{C0C0C0}96.11 \\ \bottomrule
\end{tabular}
}
\caption{MNIST, FC1, k=$5\%$}
\end{subtable}

\begin{subtable}[h]{\linewidth}
\centering
\resizebox{0.7\linewidth}{!}{%
\begin{tabular}{@{}cllllllllll@{}}
\toprule
\multicolumn{1}{l}{}          & class 0                       & class 1                       & class 2                       & class 3                       & class 4                       & class 5                       & class 6                       & class 7                       & class 8                       & class 9                       \\ \midrule
\multicolumn{1}{c|}{module 0} & \cellcolor[HTML]{C0C0C0}98.76 & 44.46                         & 44.75                         & 44.53                         & 44.53                         & 45.07                         & 44.75                         & 44.92                         & 44.55                         & 44.83                         \\
\multicolumn{1}{c|}{module 1} & 43.69                         & \cellcolor[HTML]{C0C0C0}99.60 & 43.79                         & 43.72                         & 43.68                         & 43.84                         & 43.80                         & 43.83                         & 43.73                         & 43.98                         \\
\multicolumn{1}{c|}{module 2} & 44.49                         & 44.37                         & \cellcolor[HTML]{C0C0C0}97.31 & 44.79                         & 44.67                         & 44.59                         & 44.51                         & 44.98                         & 44.75                         & 44.43                         \\
\multicolumn{1}{c|}{module 3} & 44.64                         & 44.79                         & 45.79                         & \cellcolor[HTML]{C0C0C0}95.22 & 44.63                         & 44.93                         & 44.64                         & 44.72                         & 45.35                         & 44.79                         \\
\multicolumn{1}{c|}{module 4} & 44.49                         & 44.36                         & 44.48                         & 44.54                         & \cellcolor[HTML]{C0C0C0}98.31 & 44.54                         & 44.97                         & 44.76                         & 44.80                         & 45.37                         \\
\multicolumn{1}{c|}{module 5} & 45.34                         & 44.89                         & 45.02                         & 45.79                         & 45.01                         & \cellcolor[HTML]{C0C0C0}96.52 & 46.56                         & 44.93                         & 45.19                         & 45.28                         \\
\multicolumn{1}{c|}{module 6} & 44.98                         & 44.71                         & 45.04                         & 44.72                         & 45.06                         & 45.41                         & \cellcolor[HTML]{C0C0C0}98.48 & 44.71                         & 44.82                         & 44.71                         \\
\multicolumn{1}{c|}{module 7} & 44.68                         & 44.71                         & 44.89                         & 44.84                         & 44.77                         & 44.77                         & 44.69                         & \cellcolor[HTML]{C0C0C0}95.61 & 44.73                         & 44.90                         \\
\multicolumn{1}{c|}{module 8} & 45.50                         & 45.66                         & 45.59                         & 45.53                         & 45.55                         & 45.87                         & 45.52                         & 45.49                         & \cellcolor[HTML]{C0C0C0}91.89 & 45.54                         \\
\multicolumn{1}{c|}{module 9} & 44.76                         & 44.33                         & 44.39                         & 44.45                         & 45.98                         & 44.81                         & 44.46                         & 45.54                         & 44.64                         & \cellcolor[HTML]{C0C0C0}96.67 \\ \bottomrule
\end{tabular}
}
\caption{MNIST, FC1, k=$10\%$}
\end{subtable}

\begin{subtable}[h]{\linewidth}
\centering
\resizebox{0.7\linewidth}{!}{%
\begin{tabular}{@{}cllllllllll@{}}
\toprule
\multicolumn{1}{l}{}          & class 0                       & class 1                       & class 2                       & class 3                       & class 4                       & class 5                       & class 6                       & class 7                       & class 8                       & class 9                       \\ \midrule
\multicolumn{1}{c|}{module 0} & \cellcolor[HTML]{C0C0C0}98.72 & 44.48                         & 44.77                         & 44.54                         & 44.54                         & 44.68                         & 44.77                         & 45.12                         & 44.69                         & 44.88                         \\
\multicolumn{1}{c|}{module 1} & 43.73                         & \cellcolor[HTML]{C0C0C0}99.24 & 43.98                         & 43.83                         & 43.72                         & 43.79                         & 44.02                         & 43.74                         & 43.86                         & 43.80                         \\
\multicolumn{1}{c|}{module 2} & 44.37                         & 44.41                         & \cellcolor[HTML]{C0C0C0}96.97 & 44.83                         & 44.44                         & 44.62                         & 44.51                         & 44.81                         & 44.52                         & 44.31                         \\
\multicolumn{1}{c|}{module 3} & 44.82                         & 44.80                         & 45.25                         & \cellcolor[HTML]{C0C0C0}94.46 & 44.92                         & 45.49                         & 44.82                         & 45.06                         & 45.16                         & 44.87                         \\
\multicolumn{1}{c|}{module 4} & 44.56                         & 44.41                         & 44.51                         & 44.58                         & \cellcolor[HTML]{C0C0C0}97.66 & 44.88                         & 44.70                         & 44.75                         & 44.84                         & 45.36                         \\
\multicolumn{1}{c|}{module 5} & 45.41                         & 45.28                         & 45.32                         & 45.47                         & 45.39                         & \cellcolor[HTML]{C0C0C0}94.76 & 46.47                         & 45.33                         & 45.74                         & 45.35                         \\
\multicolumn{1}{c|}{module 6} & 45.11                         & 44.69                         & 44.93                         & 44.73                         & 44.94                         & 45.10                         & \cellcolor[HTML]{C0C0C0}98.08 & 44.83                         & 44.76                         & 44.72                         \\
\multicolumn{1}{c|}{module 7} & 44.34                         & 44.23                         & 44.74                         & 44.71                         & 44.52                         & 44.71                         & 44.26                         & \cellcolor[HTML]{C0C0C0}97.53 & 44.31                         & 45.09                         \\
\multicolumn{1}{c|}{module 8} & 45.38                         & 45.33                         & 45.58                         & 45.39                         & 45.23                         & 45.97                         & 45.45                         & 45.24                         & \cellcolor[HTML]{C0C0C0}93.18 & 45.21                         \\
\multicolumn{1}{c|}{module 9} & 44.59                         & 44.39                         & 44.46                         & 44.95                         & 45.36                         & 44.68                         & 44.55                         & 45.26                         & 44.79                         & \cellcolor[HTML]{C0C0C0}96.49 \\ \bottomrule
\end{tabular}
}
\caption{MNIST, FC1, k=$20\%$}
\end{subtable}

\caption{Classification accuracy for each class for each module. The results colored in gray indicate the accuracy when the module and the subtask to be classified match. Each module holds only the classification function corresponding to its own class, and cannot classify other classes. This result indicates that each module retains only the ability to classify a single class.}
\label{tab:mono}
\end{table*}

\end{document}